\documentclass[conference]{IEEEtran}

\usepackage[utf8]{inputenc}
\usepackage[T1]{fontenc}
\usepackage[main=english]{babel}
\usepackage{microtype}
\usepackage{csquotes}

\usepackage[table]{xcolor}
\usepackage{amsmath}
\usepackage{amssymb}
\usepackage{graphicx}
\usepackage{booktabs}
\usepackage{array}
\usepackage{tabularx}
\usepackage{ragged2e}
\usepackage{enumitem}
\usepackage{adjustbox}
\usepackage{multirow}
\usepackage{tikz}
\usepackage{xurl}
\usepackage[hidelinks]{hyperref}
\usepackage{balance}
\usepackage[numbers,square,sort&compress]{natbib}
\newcommand{\parencite}[1]{\citep{#1}}
\usetikzlibrary{positioning, arrows.meta, shapes.geometric, calc, fit, backgrounds}

\newcolumntype{C}[1]{>{\centering\arraybackslash}m{#1}}
\newcolumntype{L}[1]{>{\RaggedRight\arraybackslash}p{#1}}
\newcolumntype{Y}{>{\RaggedRight\arraybackslash}X}

\hypersetup{
  pdftitle={AI Exposure and AI Resilience A Two-Dimensional Assessment Framework for Software and Software-Based Business Models},
  pdfauthor={Paul Darius Mandl, Peter Mandl, Martin Häusl},
  pdfkeywords={AI exposure, AI resilience, AI-ER, tech due diligence, generative AI, business model assessment}
}

\title{AI Exposure and AI Resilience\\
A Two-Dimensional Assessment Framework for Software and Software-Based Business Models}

\author{
\IEEEauthorblockN{Paul Darius Mandl}
\IEEEauthorblockA{Findustrial GmbH\\
Schörfling am Attersee, Austria\\
paul.mandl@findustrial.io}
\and
\IEEEauthorblockN{Peter Mandl}
\IEEEauthorblockA{Munich University of Applied Sciences\\
Munich, Germany\\
peter.mandl@hm.edu}
\and
\IEEEauthorblockN{Martin Häusl}
\IEEEauthorblockA{Munich University of Applied Sciences\\
Munich, Germany\\
martin.haeusl@hm.edu}
}

\begin{document}

\maketitle

\begin{abstract}
Artificial intelligence is changing both software production and the economics of
software-based business models. Classical technology due diligence mainly examines
technical properties such as architecture, scalability, and technical debt. These
criteria do not fully capture how AI can affect a company's value proposition,
competitive position, margins, or access to customers. This paper develops
\emph{Artificial Intelligence Exposure and Resilience} (AI-ER) as a two-dimensional
assessment framework. \emph{AI exposure} describes the pressure for change that AI
creates for a business model. \emph{AI resilience} describes the company's ability to
absorb that pressure, adapt to changed conditions, and use AI in an economically viable
way. Metrics for both dimensions are derived from current AI capabilities, their
deployment conditions, and relevant research on business models and organizational
adaptability. The model keeps exposure and resilience separate and adds an explicit
assessment of evidence quality and confidence. It can be applied first with public
information and later refined with internal evidence. The result is a traceable company
profile that supports comparison without concealing uncertainty in the underlying
evidence. The paper also specifies an initial score logic and a procedure for empirical
validation.
\end{abstract}

\begin{IEEEkeywords}
AI exposure, AI resilience, AI-ER, generative AI, software business models, tech due diligence, M\&A, AI unit economics
\end{IEEEkeywords}

\section{Introduction}
\label{sec:einleitung}

Advances in artificial intelligence are changing how companies create value and how
software-based business models should be assessed. The relevant question is no longer
only whether a company uses AI. It is also necessary to examine which parts of its
offering come under pressure and how well the company can respond. This is particularly
important for software businesses because their customer benefit often depends on the
processing or generation of information. Current AI systems can perform an increasing
share of such work at growing scale and falling unit cost
\parencite{Kanbach2024,OECDGenAI}.

The resulting assessment problem is neither purely technical nor purely strategic.
Technology due diligence typically examines architecture, code quality, scalability,
security, and technical debt. Strategic analysis focuses more strongly on customer
value, competitive position, and the economic logic of the business. AI can affect both
sides at the same time. A company may use AI extensively and still face substantial
external pressure. Conversely, proprietary data, strong customer relationships, or
regulatory barriers may provide resilience even when visible AI functionality is still
limited.

This paper develops \emph{Artificial Intelligence Exposure and Resilience} (AI-ER) to
separate these two questions. AI exposure captures the pressure that AI places on the
business model, its value proposition, margins, and customer access. AI resilience
captures the technical, organizational, and economic conditions that allow a company to
absorb this pressure and adapt. The two dimensions are assessed separately and shown
together in a two-dimensional company profile. They are not combined into one overall
score.

The paper makes three contributions. First, it defines AI exposure and AI resilience as
distinct assessment dimensions. Second, it derives a compact set of metrics from
economic impact mechanisms and specifies a non-compensatory score logic. The score logic
is complemented by an evidence and confidence model so that weak support for a rating
remains visible. Third, the paper provides a numerical example and proposes an empirical
validation procedure based on independent assessments, reliability analysis, and
sensitivity testing.

The framework draws on the capabilities and deployment limits of current AI systems as
well as research on technological exposure, business model change, platform economics,
organizational resilience, and software delivery. These foundations are used to derive
the impact mechanisms and metrics. The resulting approach is intended for strategic
assessment, investment and acquisition decisions, and technology due diligence. Its
purpose is not to produce a universal company grade. It provides a structured view of
AI-related pressure, response capacity, and the reliability of the evidence used for
both.

\section{Methodological Approach}
\label{sec:methodik}
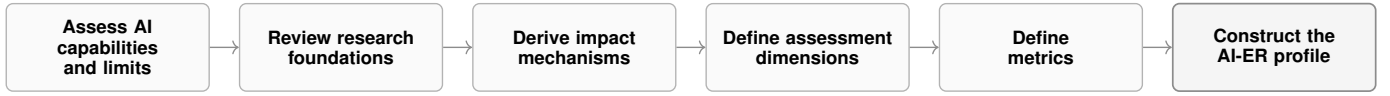
\begin{figure*}[!t]
\centering
\resizebox{\textwidth}{!}{%
\begin{tikzpicture}[
font=\footnotesize\sffamily,
box/.style={
draw=black!30,
line width=0.6pt,
fill=black!2,
rounded corners=3pt,
align=center,
text width=2.85cm,
minimum width=3.05cm,
minimum height=1.35cm,
inner sep=4pt
},
finalbox/.style={
box,
draw=black!45,
line width=0.75pt,
fill=black!4
},
arrow/.style={
->,
line width=0.6pt,
draw=black!45
}
]

\node[box] (cap) at (0,0) {%
{\bfseries Assess AI capabilities\\and limits}
};

\node[box] (lit) at (3.6,0) {%
{\bfseries Review research\\foundations}
};

\node[box] (mech) at (7.2,0) {%
{\bfseries Derive impact\\mechanisms}
};

\node[box] (dim) at (10.8,0) {%
{\bfseries Define assessment\\dimensions}
};

\node[box] (metrics) at (14.4,0) {%
{\bfseries Define\\metrics}
};

\node[finalbox] (profile) at (18.0,0) {%
{\bfseries Construct the\\AI-ER profile}
};

\draw[arrow] (cap.east) -- (lit.west);
\draw[arrow] (lit.east) -- (mech.west);
\draw[arrow] (mech.east) -- (dim.west);
\draw[arrow] (dim.east) -- (metrics.west);
\draw[arrow] (metrics.east) -- (profile.west);

\end{tikzpicture}%
}

\caption{Procedure for developing the AI-ER assessment framework, from the conceptual foundations to the two-dimensional company profile.}
\label{fig:methodik}
\end{figure*}

The framework translates two abstract concepts into assessment questions that can be
examined against observable information. AI exposure represents external pressure for
change. AI resilience represents the capacity to respond. A purely verbal appraisal
would make comparisons across companies, assessors, and points in time difficult.
AI-ER therefore uses defined metrics for both dimensions.

The derivation combines a technical and an economic perspective. Current AI capabilities
indicate which services can in principle be provided by AI and which deployment
conditions limit practical use. Research on business models and market structure shows
when these capabilities can create economic pressure. Research on organizational
adaptability identifies the conditions that allow a company to respond. The framework
links these perspectives without inferring a company's exposure directly from a single
technical capability.

Recurring economic impact mechanisms are derived first. They describe how AI can alter a
company's offering, market position, or economic performance. The mechanisms are then
assigned to AI exposure or AI resilience and translated into candidate metrics. A metric
is retained when it has a distinct conceptual role, can be supported by observable
evidence, and does not duplicate another assessment quantity.

The number of metrics is deliberately limited. Too many indicators would increase
collection effort and make overlaps more likely. Excessive aggregation would conceal
important differences between business model, market position, and organizational or
technical capability. The final selection therefore follows from the substantive
derivation in Chapter~\ref{sec:herleitung}. Figure~\ref{fig:methodik} summarizes this
procedure.

Metric scores are combined only within their respective dimension. The dimension scores
make the main tendency easier to compare, while the individual metrics remain visible.
This distinction matters because high exposure and high resilience can occur at the same
time. The company profile therefore depends on the joint interpretation of both
dimensions rather than on one aggregate score.

Each metric rating is linked to documented evidence. Confidence describes how well a
rating is supported rather than how high the rating is. It depends on the quality of the
available evidence and, where several independent assessment runs exist, on the agreement
between those runs. Chapter~\ref{sec:umsetzung} specifies the rating, aggregation, and
confidence logic.
\section{Capabilities and Deployment Conditions of Artificial Intelligence}
\label{sec:ki-faehigkeitsprofile}

AI-ER requires a technical foundation that is not tied to one model, vendor, or product
generation. The relevant issue is therefore not which specific AI technology a company
uses. The framework asks which economically relevant services AI systems can perform
under operational conditions. This functional view connects technical development to
possible changes in a business model without treating the availability of a technical
feature as evidence of economic impact.

\subsection{Economically Relevant Capability Areas}

Current AI systems provide analytical, predictive, generative, and decision-support
services. Analytical methods identify patterns and anomalies in large data sets and can
support the automated evaluation of complex information. Their performance is documented
in fields such as image classification and medical image analysis
\parencite{AIIndex2026}. Predictive systems estimate probabilities and expected
developments from existing data. Their usefulness depends strongly on data quality and
on the stability of the relationships learned from those data. Optimization methods
extend these capabilities by comparing alternative courses of action within defined
objectives and constraints \parencite{AIIndex2026}.

Generative AI adds the creation and transformation of content. Current systems can
produce text, images, and program code and can transform information between different
representations. Standardized evaluations and professional task benchmarks document the
breadth of these capabilities \parencite{AIIndex2026,IASafetyReport2026}. For business
use, however, the quality of a single output is not sufficient. Economic relevance
increases when a service can be delivered repeatedly, integrated into existing systems,
and scaled at acceptable cost. Interfaces and standardized services make such
capabilities accessible even to companies that do not develop foundation models
themselves.

This broad access has an important economic consequence. Similar AI capabilities may be
available to a company, its competitors, its customers, and large platform providers at
the same time. Technical access therefore does not in itself create a durable advantage.
The later assessment must examine where AI becomes economically effective and which
company-specific conditions support or limit its use.

\subsection{Deployment Conditions and Limits}

The documented capabilities do not make AI a universal replacement technology.
Performance remains dependent on the task, the available data, and the operating
context. Results from standardized evaluations cannot simply be transferred to business
processes. Generative systems can produce plausible but incorrect output and can react
unstably to changed inputs. They may also fail to reflect company-specific rules or
exceptions. These limitations are particularly relevant in complex or
liability-sensitive applications \parencite{IASafetyReport2026,NISTGenAIProfile2024}.

Further limits arise when isolated AI functions are embedded in longer workflows.
Operational use requires state to be maintained, errors to be detected, and
responsibilities to remain clear across several processing steps. Uncertainty can
accumulate as workflows become longer. A deployable solution therefore requires more
than a capable model. It also needs integration, monitoring, and suitable human control.

Technical feasibility also differs from economic viability. Operation, quality
assurance, and integration create costs that can reduce the achievable benefit. Legal
and organizational requirements may further restrict use. Data protection, information
security, traceability, and human oversight are especially relevant where decisions
concern persons or protected information
\parencite{NISTGenAIProfile2024,EUAIAct2024}. AI-ER therefore treats a capability as
economically relevant only when it can be used with sufficient reliability,
availability, viability, and permissibility in the respective context.

The technical analysis defines the range of services from which AI-related change may
originate. It does not yet establish whether a particular company is exposed or
resilient. That assessment also depends on economic and organizational relationships,
which are considered in the next chapter.

\section{State of Research and Conceptual Frame}
\label{sec:stand-wissenschaft}

Technical capability becomes economically relevant when it changes services, cost
structures, market relationships, or forms of value creation. AI-ER therefore combines
three research perspectives. Work on technological exposure shows where AI capabilities
overlap with existing activities and services. Business model research explains how
such overlaps can affect value creation and economic capture. Research on resilience
addresses how organizations respond to change. These perspectives provide the
conceptual basis for the two assessment dimensions.

\subsection{Technological Exposure and Business Model Change}
\label{sec:stand-exposure}

Research on technological exposure examines how strongly activities or occupations
overlap with the capabilities of a new technology. For generative AI, Eloundou et al.\
and the OECD identify particularly strong exposure for knowledge-intensive,
language-based, and information-processing activities
\parencite{Eloundou2024,OECDExposure2026}. Such proximity indicates technical potential
for change. It does not establish that an activity will be automated or that a company
will be economically weakened. Operational and market conditions still determine
whether the technology can be used and how its effects are distributed.

A company assessment must therefore extend beyond isolated activities. The relevant
question is whether AI changes the economic core of a service, its value proposition, or
its market position. Research on business model innovation shows that AI can support new
offerings, alter service delivery, and change economic capture
\parencite{Kanbach2024,Jorzik2024}. The effect depends on the business logic. A digital
information product is generally closer to generative and analytical AI capabilities
than an offering whose value depends heavily on physical, regulatory, or relational
conditions.

Platform economics adds the question of market structure and customer access. Digital
platforms can integrate formerly independent functions into larger offerings and can
take over the interface to the customer \parencite{Parker2016,Wessel2025}. A service may
therefore remain useful while its independent monetization or route to market weakens.
AI exposure in this paper consequently includes more than technical substitutability. It
also covers AI-related changes in value creation, competition, and customer access.

\subsection{Organizational Resilience and Adaptability}
\label{sec:stand-resilienz}

Organizational resilience concerns the ability to deal with change and disruption. It
includes anticipation, continued capacity to act, and subsequent adaptation. Duchek
describes resilience as a capability that links anticipation, coping, and adaptation
\parencite{Duchek2020}. This perspective is well suited to AI-related change because new
capabilities diffuse over time and can repeatedly alter customer expectations and
competitive conditions.

Research on digital resilience applies this perspective to organizations whose
operations depend strongly on digital technologies. Digital technologies can improve
information processing and responsiveness, but they can also create dependencies on
data, platforms, infrastructure, and external providers
\parencite{TimLeidner2023}. The use of AI therefore does not automatically increase
resilience. Technical possibilities must be converted into dependable operational
capabilities and linked to organizational decisions.

Business model research also treats adaptation as a response to environmental change.
Buliga et al.\ describe business model change as one such response
\parencite{Buliga2016}. More recent studies examine AI as a resource that may support
organizational resilience. Han et al.\ find that AI investment can help firms respond to
external disruptions when complementary organizational conditions are present
\parencite{Han2025}. Guo et al.\ report a positive relationship between AI use and
organizational resilience that is partly mediated by business model development
\parencite{Guo2026}. AI-ER changes the direction of the question. It asks how well a
company can respond to change that is itself induced by AI.

For the framework, AI resilience therefore denotes the ability to absorb AI-related
pressure and to adapt products, processes, and the business model where necessary. This
capacity may rest on existing competitive positions as well as technical and
organizational capabilities. It is not the inverse of exposure. High exposure can
coexist with high resilience, while low exposure can coexist with weak adaptability.

\subsection{Research Gap and Conceptual Delineation}
\label{sec:forschungsluecke}

Existing research covers important parts of AI-related change but rarely combines them
for the assessment of an individual company. Exposure studies often focus on tasks or
occupations. Business model research examines changes in value creation but does not
always connect them to concrete proximity to AI capabilities. Resilience research
usually starts from general environmental change and increasingly treats AI as a
supporting resource.

AI-ER connects these perspectives. AI exposure denotes the pressure that AI capabilities
and their diffusion place on a business model and its market position. AI resilience
denotes the company's capacity to absorb this pressure, adapt, and use new technical
possibilities for its own value creation. Keeping the dimensions separate prevents
technical vulnerability and organizational response capacity from being reduced to one
indicator.

The research base does not determine a unique set of metrics. It defines the areas that
the assessment needs to cover. For exposure, these areas concern the transfer of AI
capabilities into value creation, competition, and market relationships. For resilience,
they concern protection, adaptability, and the productive use of AI. The next chapter
derives a compact metric set from these foundations.

\section{Derivation of the AI-ER Assessment Framework}
\label{sec:herleitung}

The preceding chapters provide the technical and conceptual basis for AI-ER. The next
step is to convert these foundations into assessment quantities. The derivation starts
with economic impact mechanisms and then assigns metrics to AI exposure and AI
resilience. Closely related aspects are merged when separate treatment would create
double counting. A metric is retained only when it adds a distinct assessment question
and can be supported by observable evidence.

\subsection{Economic Impact Mechanisms}
\label{sec:mechanismen}

An AI capability becomes economically relevant when it changes how a service is
provided, differentiated, or monetized. Research on business model innovation and
platform economics points to four recurring mechanisms
\parencite{Kanbach2024,Parker2016,Wessel2025}.

\begin{itemize}

\item \emph{Substitution} occurs when AI can provide a substantial part of the customer
benefit for which the existing service is paid. Comparable quality at lower effort or
shorter delivery time can already create pressure.

\item \emph{Compression} reduces the economic value contribution without eliminating
the service. AI may lower labor input or standardize previously scarce expertise. This
can reduce prices and margins.

\item \emph{Bundling} integrates a formerly stand-alone service into a broader product
or platform. The function remains available, but its independent economic position can
weaken.

\item \emph{Re-intermediation} changes the relationship between provider and customer.
Assistants, platforms, or agent systems may take over search, selection, or transaction
steps and thereby affect customer access.

\end{itemize}

Several mechanisms can occur at the same time. They are therefore treated as forms of
economic impact rather than as mutually exclusive development paths.

\subsection{Deriving the AI Exposure Metrics}
\label{sec:exposure-herleitung}

AI exposure measures the pressure that these mechanisms place on a company. Research on
task exposure first motivates the \emph{substitutability of the core benefit}
\parencite{Eloundou2024,OECDExposure2026}. This metric asks how closely the paid customer
benefit lies to services that current AI systems can already provide. The focus is on
the service itself rather than on the current strength of brand, sales, or customer
relationships.

A second metric captures the \emph{replicability of the offering}. Generative AI can
reduce the effort required to build or reproduce parts of digital products. Replicability
therefore considers the effort needed to recreate the offering while taking account of
company-specific data, integrations, domain knowledge, and regulatory requirements.
Substitutability and replicability remain separate because a difficult-to-copy product
may still lose its customer benefit to a different AI-based solution. The reverse is
also possible.

Technical replaceability alone does not determine market pressure. \emph{Competitive
dynamics} captures how competitors or platform providers use AI to change speed, price,
or bundling. The \emph{business model modulator} captures the direction of the economic
effect on the existing business model. AI-related productivity or quality gains can
dampen exposure when the company can retain the resulting value. The same developments
can amplify exposure when they mainly benefit customers, competitors, or new entrants.

\emph{Customer access and demand pressure} covers changes on the demand side. Customers
may expect AI functions, perform parts of a service themselves, or delegate selection
to assistants and platforms \parencite{Bick2025,OECDCompetition2025}. This can alter
willingness to pay and the direct relationship between provider and customer. The metric
is therefore distinct from competitive dynamics, which focuses on the behavior of other
providers.

The derivation yields four numerical exposure metrics and one categorical business model
modulator. Together they cover the core benefit, the replicability of the offering,
competitive change, the direction of the business model effect, and customer access.
This scope is broad enough to represent the main impact mechanisms while remaining
manageable for evidence collection.

\subsection{Deriving the AI Resilience Metrics}
\label{sec:resilienz-herleitung}

AI resilience describes the company's capacity to respond to AI-related pressure.
Business model and platform research first points to \emph{protective positions}. These
are company-specific resources or market positions that remain valuable under changed
technical conditions. The metric assesses their robustness against AI-based alternatives
rather than their historical strength alone.

Research on organizational resilience and dynamic capabilities motivates
\emph{adaptability} \parencite{Duchek2020,Teece1997}. The metric concerns the ability to
recognize relevant change and to adjust products, processes, and the business model. It
therefore describes an organizational capability rather than the success of a single AI
project.

Productive AI use also requires a technical basis. \emph{Technical AI maturity}
assesses whether data-driven and model-based functions can be developed, integrated,
monitored, and operated reliably
\parencite{Kreuzberger2023,Sculley2015}. \emph{Implementation capability} addresses a
different question. It examines whether skills, decision paths, and software delivery
allow technical possibilities to be converted into operational solutions within a
reasonable period \parencite{Forsgren2018}.

The final resilience metric is \emph{economic viability}. It compares the effort for
development, operation, control, and external dependencies with the expected
contribution to value creation and customer benefit. This prevents technical feasibility
from being treated as sufficient evidence of lasting economic value.

The five resilience metrics cover protective positions, adaptability, technical AI
maturity, implementation capability, and economic viability. They describe distinct
conditions that influence a company's ability to respond. The next chapter specifies
how the exposure and resilience metrics are rated and combined.

\subsection{Overview of the Metrics}
\label{sec:metriken-uebersicht}

The derivation yields four numerical exposure metrics, one categorical business model
modulator, and five numerical resilience metrics. Table~\ref{tab:metriken} summarizes
their conceptual basis.

\begin{table*}[!t]
\centering
\footnotesize
\caption{Overview of the assessment quantities derived in the AI-ER framework and of
their conceptual derivation.}
\label{tab:metriken}
\vspace{4pt}

{\renewcommand{\arraystretch}{1.18}
\begin{tabularx}{\textwidth}{
  p{2.0cm}
  >{\raggedright\arraybackslash}p{4.1cm}
  X
  >{\raggedright\arraybackslash}p{1.6cm}
}
\toprule
\textbf{Dimension}
& \textbf{Assessment quantity}
& \textbf{Subject of assessment}
& \textbf{Derived from} \\
\midrule

AI exposure
& Substitutability of the core benefit
& Takeover of the essential customer benefit by AI
& \parencite{Eloundou2024,OECDExposure2026} \\

& Replicability of the offering
& Effort required to reproduce the offering
& \parencite{Eloundou2024,Kanbach2024} \\

& Competitive dynamics
& AI-induced changes in competition, prices, and bundling
& \parencite{Kanbach2024,Parker2016,Wessel2025} \\

& Business model modulator
& Dampening, neutral, or amplifying effect of AI on the economic position
& \parencite{Kanbach2024,Parker2016} \\

& Customer access and demand pressure
& Changes in customer expectations, demand, and market access
& \parencite{Bick2025,OECDCompetition2025,Wessel2025} \\

\midrule

AI resilience
& Protective positions
& Robust resources and market positions
& \parencite{Parker2016,Teece1997} \\

& Adaptability
& Further development of products, processes, and business model
& \parencite{Duchek2020,Teece1997} \\

& Technical AI maturity
& Development, integration, and operation of productive AI systems
& \parencite{Kreuzberger2023,Sculley2015} \\

& Implementation capability
& Conversion of technical possibilities into operational solutions
& \parencite{Forsgren2018,Teece1997} \\

& Economic viability
& Relation of effort, benefit, and lasting economic contribution
& \parencite{Sculley2015,Kreuzberger2023} \\

\bottomrule
\end{tabularx}
}
\end{table*}

The table shows that each assessment quantity addresses a distinct aspect of AI-related
change. The following chapter defines the rating logic and the treatment of evidence and
confidence.
\section{Proposal for a Concrete Implementation}
\label{sec:umsetzung}

The derived metrics require an explicit rating and aggregation procedure before they can
be used in practice. Different implementations are possible, including averaging and
multi-criteria procedures. This paper specifies one rule-based variant. The purpose of
the proposal is to keep critical individual scores visible and to provide a configuration
that can be tested empirically.

AI-ER does not produce a general company grade. It classifies AI-related pressure and a
company's capacity to respond. The same metrics can be used for an initial assessment
based on public information and for a later assessment with internal evidence. A richer
evidence base may change metric scores, dimension scores, and confidence values without
changing the underlying model.

An assessment can be performed by one assessor, by several independent assessors, or by
an AI-based analysis service. An automated service may collect information about a
company, its offering, and relevant competitors and then apply the defined rating rules.
Reliability improves when several assessment runs are produced independently and compared
afterward. Runs are considered independent only when they do not share intermediate
judgments. Repeated outputs from the same reasoning process do not become independent
merely because several outputs are generated.

The formal specification makes the rating logic reproducible and provides a basis for
software implementations of AI-ER. Such implementations can collect evidence, propose
metric scores, execute assessment runs, and document disagreements.

\subsection{Notation}
\label{subsec:notation}

Only the terms needed for the score logic are introduced here. The appendix provides the
complete mathematical notation.

\begin{itemize}[
leftmargin=1.5em,
label=\textbullet,
itemsep=0.55\baselineskip,
topsep=0.4\baselineskip,
parsep=0pt
]

\item
A \textbf{\emph{metric}} is a company characteristic rated against defined scale
anchors. Its value is the \textbf{\emph{metric score}}. Metric scores are combined into
a \textbf{\emph{dimension score}} according to the specified score logic.

\item
A \textbf{\emph{scale anchor}} assigns a substantive meaning to a value on the
five-point rating scale. The values 1, 3, and 5 are described explicitly. The values 2
and 4 represent justified intermediate positions.

\item
\textbf{\emph{Core drivers}} determine the base value of a dimension. The proposed
logic is \textbf{\emph{non-compensatory}}, which means that a critical core driver is not
automatically offset by a favorable value on another core driver.

\item
A \textbf{\emph{threshold}} activates a predefined rule once a specified value is
reached. A \textbf{\emph{combination rule}} defines how scores or conditions are linked.
A \textbf{\emph{binary indicator}} records whether such a condition is fulfilled.

\item
A \textbf{\emph{modulator}} changes a base value by a bounded amount when an additional
economic or structural condition has a dampening or amplifying effect. It does not alter
the underlying metric ratings.

\item
An \textbf{\emph{assessment run}} is one complete application of the rating scheme to a
company. Several independent runs can be combined into a joint metric score and used to
measure rating agreement.

\end{itemize}

Thresholds, combination rules, and modulators are configurable parameters rather than
values inferred from company data. They must be fixed before an assessment is applied.
The configuration proposed below is an initial specification and requires empirical
comparison with alternative settings. Table~\ref{tab:notation} summarizes the symbols.
Table~\ref{tab:parameter} lists the configurable model parameters and the default values
used in this paper.

\subsection{Determining the Individual Scores and the Score Logic}
\label{sec:score-logik}

Substantive scale anchors are defined for each numerical metric before the assessment,
describing observable conditions for low, medium, and high levels. The values 1, 3, and
5 come directly from these anchors, while 2 and 4 represent justified intermediate
positions. Missing evidence does not automatically produce a medium score. The
assessment is marked provisional or suspended until sufficient evidence becomes
available. For each metric, the assigned score, its justification, and the supporting
evidence are documented together.

For company $i$, metric $j$, and assessment run $\ell$,
$x_{ij}^{(\ell)}\in\{1,2,3,4,5\}$. If $L\geq 2$ independent ratings are available, the
median is used as the robust joint metric score.

\begin{equation}
\widetilde{x}_{ij}
=
\operatorname{median}\!\left(
  x_{ij}^{(1)},\ldots,x_{ij}^{(L)}
\right).
\label{eq:metric-median}
\end{equation}

The median limits the influence of strongly deviating ratings without assuming equal
distances between the five scale levels. The default configuration for the initial
outside-in assessment uses $L=3$. The subsequent inside-in assessment described in
Section~\ref{sec:analyseschichten} uses a single run. The median therefore remains an
observed integer scale value when several independent runs are used. If an even number
of runs is used, the rule for selecting one of the two middle values must be specified
in advance. With a single run, $\widetilde{x}_{ij}=x_{ij}^{(1)}$. The symbols
$x_i^{\mathrm{sub}}$ through $x_i^{\mathrm{econ}}$ denote the resulting combined metric
scores.

The business model modulator is treated separately because it does not describe an
ordinal intensity. It records whether AI dampens, leaves unchanged, or amplifies the
pressure associated with substitutability and replicability. The classification depends
on the economic effect on the company rather than on technical AI usability alone.
Productivity or quality gains matter only to the extent that they change the company's
position relative to customers, competitors, and possible substitutes.

A dampening effect is present when AI strengthens the economic position of the existing
business model. This may occur when the company captures AI-induced gains while retaining
important competitive advantages. Proprietary data, durable customer relationships, or
regulatory requirements can support such an effect. A neutral effect is assigned when
neither direction predominates or when the evidence does not permit a reliable
classification. An amplifying effect is present when AI weakens the business model, for
example by making services easier to substitute or by shifting value toward customers,
competitors, or new entrants.

With several assessment runs, the business model modulator is classified independently
in each run and consolidated afterward. Diverging assignments are documented and checked
against the evidence. The consolidated value enters the exposure score directly as
$M_i^{\mathrm{bm}}$. It is not averaged with the numerical metric scores.

The two dimension scores use non-compensatory aggregation, a principle established in
multi-criteria decision analysis \parencite{Roy1996,Munda2008}. A simple average could
hide a critical attack path or a substantial weakness behind favorable scores on other
metrics. The proposed logic therefore uses core drivers, binary indicators, and
modulators. The indicator function is defined below.

\begin{equation}
I(P)
=
\begin{cases}
1, & \text{if $P$ holds},\\
0, & \text{otherwise}.
\end{cases}
\label{eq:indicator}
\end{equation}

For AI exposure, substitutability of the core benefit $x_i^{\mathrm{sub}}$ and
replicability of the offering $x_i^{\mathrm{rep}}$ are the core drivers. Either can
create substantial pressure on its own. A product may be difficult to reproduce while
its customer benefit is replaced by another AI-based solution. Conversely, high
replicability can increase competitive pressure before the core benefit is fully
substituted. The base value is therefore
$\max\{x_i^{\mathrm{sub}},\,x_i^{\mathrm{rep}}\}$. The maximum preserves the more
critical of the two attack paths.

Competitive dynamics $x_i^{\mathrm{comp}}$ and customer access and demand pressure
$x_i^{\mathrm{cust}}$ jointly enter an additional indicator. In the proposed
configuration, a score of at least 4 activates the indicator.

\begin{equation}
I_i^E
=
I\!\left(
  x_i^{\mathrm{comp}}\geq 4
  \;\lor\;
  x_i^{\mathrm{cust}}\geq 4
\right).
\label{eq:exposure-indicator}
\end{equation}

The categorically rated business model modulator is represented as follows.

\begin{equation}
M_i^{\mathrm{bm}}
=
\begin{cases}
-1, & \text{for a dampening effect},\\
0,  & \text{for a neutral or ambiguous effect},\\
1,  & \text{for an amplifying effect}.
\end{cases}
\label{eq:business-model-modulator}
\end{equation}

The exposure score is then

\begin{equation}
E_i
=
\operatorname{clamp}_{[1,5]}\!\left(
  \max\!\left(x_i^{\mathrm{sub}},x_i^{\mathrm{rep}}\right)
  + I_i^E
  + M_i^{\mathrm{bm}}
\right).
\label{eq:exposure-score}
\end{equation}

For AI resilience, protective positions $x_i^{\mathrm{prot}}$ and adaptability
$x_i^{\mathrm{adapt}}$ are the core drivers under a weakest-link logic. Strong protective
positions cannot fully compensate for low adaptability, and adaptability does not replace
missing structural protection. The base value is therefore
$\min\{x_i^{\mathrm{prot}},\,x_i^{\mathrm{adapt}}\}$. The minimum keeps the weaker core
condition visible.

Technical AI maturity $x_i^{\mathrm{tech}}$ and implementation capability
$x_i^{\mathrm{impl}}$ increase the base value only when both are high. Technical
infrastructure without implementation capability is insufficient, as is organizational
readiness without a reliable technical foundation.

\begin{equation}
I_i^R
=
I\!\left(
  x_i^{\mathrm{tech}}\geq 4
  \;\land\;
  x_i^{\mathrm{impl}}\geq 4
\right).
\label{eq:resilience-indicator}
\end{equation}

Low economic viability is captured through a separate modulator.

\begin{equation}
M_i^{\mathrm{econ}}
=
I\!\left(x_i^{\mathrm{econ}}\leq 2\right).
\label{eq:economic-modulator}
\end{equation}

The resilience score is then

\begin{equation}
R_i
=
\operatorname{clamp}_{[1,5]}\!\left(
  \min\!\left(x_i^{\mathrm{prot}},x_i^{\mathrm{adapt}}\right)
  + I_i^R
  - M_i^{\mathrm{econ}}
\right).
\label{eq:resilience-score}
\end{equation}

The function $\operatorname{clamp}$ bounds both dimension scores to the interval from 1
to 5.

\begin{equation}
\operatorname{clamp}_{[1,5]}(z)
=
\min\!\bigl(5,\max(1,z)\bigr).
\label{eq:clamp}
\end{equation}

The aggregation operators have different roles. The median combines independent
ratings robustly. The maximum preserves a sufficient attack path on the exposure side,
while the minimum keeps a non-compensable weakness visible on the resilience side.
Indicators represent threshold conditions. Modulators adjust the base value by a bounded
step when an additional contextual factor changes the direction of the assessment. The
clamping function keeps the final dimension scores on the common five-point scale.

Integer scores are intentional in the default configuration. The framework provides a
classification on five levels rather than a degree of numerical precision that the
evidence does not support. Thresholds, combination rules, and modulator values are part
of the proposed configuration and must be compared empirically with alternatives.

\subsection{Evidence, Consistency, and Confidence}
\label{sec:konfidenz}

Every metric score must be supported by observable information. The relevant evidence
depends on the metric. Each assessment records which statements are directly supported,
which depend on inference, and where information remains incomplete.

Evidence quality is assessed through five properties. \emph{Directness} indicates how
closely the evidence relates to the metric itself. \emph{Timeliness} reflects whether
the information is current enough for the assessment. \emph{Completeness} captures
whether the relevant aspects of the metric are covered. \emph{Independence} reduces the
risk of treating repeated information from dependent sources as separate confirmation.
\emph{Agreement} reflects whether independent sources support compatible conclusions.
Each property is rated 0 for insufficient, 0.5 for partly fulfilled, or 1 for fulfilled.

The five properties need not contribute equally to every metric. Positive weights can
therefore be assigned to the evidence components. The weights may differ by metric. The
default configuration uses equal weights, which provides a simple reference case for
later validation.

For company $i$, metric $j$, and evidence property $k$, let
$q_{ij}^{(k)}\in\{0,0.5,1\}$ denote the component score and let $w_j^{(k)}>0$ denote its
normalized weight. The weights satisfy $\sum_{k\in\mathcal{K}}w_j^{(k)}=1$. The set of
evidence properties is
\[
\mathcal{K}
=
\{\mathrm{dir},\mathrm{tim},\mathrm{cmp},
\mathrm{ind},\mathrm{agr}\}.
\]
Evidence quality is calculated as a weighted mean.

\begin{equation}
Q_{ij}
=
\sum_{k\in\mathcal{K}}
w_j^{(k)} q_{ij}^{(k)},
\qquad
Q_{ij}\in[0,1].
\label{eq:evidence-quality-general}
\end{equation}

For the five properties used in the framework, the formula can be written explicitly as

\begin{equation}
\begin{aligned}
Q_{ij}
={}&
w_j^{\mathrm{dir}} q_{ij}^{\mathrm{dir}}
+w_j^{\mathrm{tim}} q_{ij}^{\mathrm{tim}}
+w_j^{\mathrm{cmp}} q_{ij}^{\mathrm{cmp}}\\
&+w_j^{\mathrm{ind}} q_{ij}^{\mathrm{ind}}
+w_j^{\mathrm{agr}} q_{ij}^{\mathrm{agr}}.
\end{aligned}
\label{eq:evidence-quality}
\end{equation}

Because the normalized weights sum to one and all component scores lie in $[0,1]$,
$Q_{ij}$ also lies in $[0,1]$. A value of 0 means that all five properties are rated
insufficient. A value of 1 requires all five properties to be fully satisfied. With equal
weights, Equation~(\ref{eq:evidence-quality}) reduces to the arithmetic mean.

The separate component scores remain part of the assessment record. Two metrics can have
the same value of $Q_{ij}$ even when the supporting evidence differs materially. One
rating may be based on current but incomplete information while another may rely on
complete information from dependent sources. The aggregate value therefore supports
comparison without replacing inspection of the component scores and the documented
justification.

When several independent assessment runs are available, their rating agreement is
measured separately. The value $A_{ij}$ captures the deviation of the $L$ individual
ratings from their joint median $\widetilde{x}_{ij}$.

\begin{equation}
A_{ij}
=
1-
\frac{1}{2L}
\sum_{\ell=1}^{L}
\left|
x_{ij}^{(\ell)}-\widetilde{x}_{ij}
\right|,
\qquad
A_{ij}\in[0,1],
\quad L\geq 2.
\label{eq:rating-consistency}
\end{equation}

The normalization uses the five-point rating scale. Its range is 4 and the mean absolute
deviation from the median can be at most 2 points. The factor $2L$ therefore maps the
agreement measure to $[0,1]$. Identical ratings produce $A_{ij}=1$. Larger deviations
reduce the value. With an even number of assessment runs, $A_{ij}=0$ can occur when half
of the ratings lie at each end of the scale. With an odd number of runs, the minimum is
greater than 0.

Two forms of agreement are kept distinct. The component
$q_{ij}^{\mathrm{agr}}$ refers to agreement among evidence sources. The value $A_{ij}$
refers to agreement among independent assessment runs. The first concerns the evidence
base. The second concerns the stability of the resulting rating across assessors.

With only one assessment run, confidence is based on evidence quality alone and is marked
as not independently confirmed. With several independent runs, confidence is limited by
the weaker of evidence quality and rating agreement.

\begin{equation}
C_{ij}
=
\begin{cases}
Q_{ij},
& \text{for $L=1$},\\[2mm]
\min\!\left\{Q_{ij},A_{ij}\right\},
& \text{for $L\geq 2$}.
\end{cases}
\qquad
C_{ij}\in[0,1].
\label{eq:metric-confidence}
\end{equation}

The minimum implements a non-compensatory rule. Strong evidence cannot compensate for
poor agreement between independent runs. High agreement cannot compensate for weak
evidence. Confidence is classified as low for $C_{ij}<0.5$, medium for
$0.5\leq C_{ij}<0.75$, and high for $C_{ij}\geq 0.75$. These thresholds are part of the
proposed configuration and require empirical validation.

Dimension confidence is based only on metrics that actually affect the corresponding
dimension score. For company $i$, $\mathcal{A}_i^{\mathrm{E}}$ denotes the
decision-relevant metrics for exposure and $\mathcal{A}_i^{\mathrm{R}}$ denotes the
decision-relevant metrics for resilience. These sets include the core driver selected
by the maximum or minimum and any metric that activates a binary indicator. They also
include metrics that determine a modulator. If several metrics are equally decisive,
all of them are included.

The confidence values of the two dimension scores are

\begin{equation}
C_i^{\mathrm{E}}
=
\min_{j\in\mathcal{A}_i^{\mathrm{E}}}
C_{ij},
\qquad
C_i^{\mathrm{R}}
=
\min_{j\in\mathcal{A}_i^{\mathrm{R}}}
C_{ij}.
\label{eq:dimension-confidence}
\end{equation}

The minimum ensures that dimension confidence does not exceed the confidence of its
weakest decision-relevant basis. A high dimension score with low confidence should
therefore be treated as a result that requires further evidence. Scores close to a
threshold should also be identified because a small reassessment can change an
indicator, a modulator, or the quadrant assignment.

\subsection{The Two-Dimensional AI-ER Profile}
\label{sec:quadranten}

AI exposure and AI resilience are displayed jointly rather than offset against each
other. In the proposed configuration, scores from 1 to 3 are classified as low and
scores of 4 or 5 as high. A score of 3 lies immediately below the threshold and should be
marked as near-threshold. Table~\ref{tab:quadrant-thresholds} defines the four quadrant
assignments.

\begin{table}[!t]
\centering
\small
\caption{Score-based assignment to the quadrants of the AI-ER profile.}
\label{tab:quadrant-thresholds}
\vspace{4pt}
\begin{tabular}{lll}
\toprule
\textbf{Quadrant} & \textbf{AI exposure} & \textbf{AI resilience} \\
\midrule
Defended Niche & $E_i\leq 3$ & $R_i\geq 4$ \\
AI-Ready Compounder & $E_i\geq 4$ & $R_i\geq 4$ \\
Rebuilding Required & $E_i\leq 3$ & $R_i\leq 3$ \\
Acute Threat & $E_i\geq 4$ & $R_i\leq 3$ \\
\bottomrule
\end{tabular}
\end{table}

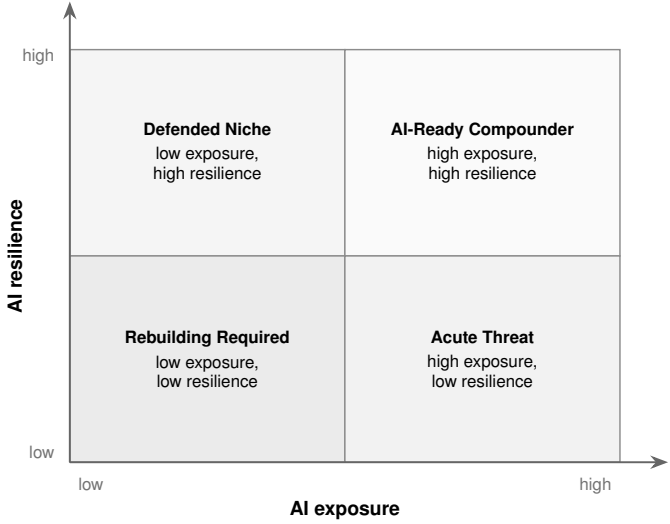
\begin{figure}[!t]
\centering
\resizebox{\columnwidth}{!}{%
\begin{tikzpicture}[font=\scriptsize\sffamily]

\filldraw[fill=black!4, draw=black!45, line width=0.6pt] (0,3) rectangle (4,6);
\filldraw[fill=black!2, draw=black!45, line width=0.6pt] (4,3) rectangle (8,6);
\filldraw[fill=black!8, draw=black!45, line width=0.6pt] (0,0) rectangle (4,3);
\filldraw[fill=black!5, draw=black!45, line width=0.6pt] (4,0) rectangle (8,3);

\draw[-{Stealth[length=2.6mm]}, line width=0.8pt, draw=black!60] (0,0) -- (8.7,0);
\draw[-{Stealth[length=2.6mm]}, line width=0.8pt, draw=black!60] (0,0) -- (0,6.7);

\node[font=\footnotesize\bfseries\sffamily] at (4,-0.7) {AI exposure};
\node[rotate=90,font=\footnotesize\bfseries\sffamily] at (-0.8,3) {AI resilience};

\node[anchor=north west, text=black!55] at (0,-0.1) {low};
\node[anchor=north east, text=black!55] at (8,-0.1) {high};
\node[anchor=east, text=black!55] at (-0.1,0.15) {low};
\node[anchor=east, text=black!55] at (-0.1,5.9) {high};

\node[align=center, font=\scriptsize\sffamily] at (2,4.5)
  {\textbf{Defended Niche}\\[2pt]
   low exposure,\\
   high resilience};

\node[align=center, font=\scriptsize\sffamily] at (6,4.5)
  {\textbf{AI-Ready Compounder}\\[2pt]
   high exposure,\\
   high resilience};

\node[align=center, font=\scriptsize\sffamily] at (2,1.5)
  {\textbf{Rebuilding Required}\\[2pt]
   low exposure,\\
   low resilience};

\node[align=center, font=\scriptsize\sffamily] at (6,1.5)
  {\textbf{Acute Threat}\\[2pt]
   high exposure,\\
   low resilience};

\end{tikzpicture}%
}
\caption{Quadrants of the two-dimensional AI-ER profile.}
\label{fig:quadrant}
\end{figure}

\begin{itemize}
  \item \emph{Defended Niche.} This position combines low exposure with high resilience.
  The core benefit is comparatively difficult to attack, while good conditions for
  adaptation and AI use are also present.

  \item \emph{AI-Ready Compounder.} High exposure meets high resilience. The company
  operates in a strongly changing environment but has the conditions required to manage
  the change actively and use it for its own development.

  \item \emph{Rebuilding Required.} Current exposure is low, but the capacity for change
  is limited. If the pressure for change rises, the company can respond only to a limited
  extent. The primary need for action therefore lies in building protective, adaptive,
  and implementation capability.

  \item \emph{Acute Threat.} High exposure coincides with low resilience. Vulnerable
  services meet insufficient protective and adaptive capability, creating an immediate
  need for review and action.
\end{itemize}

The quadrants are not final company classes. Interpretation must also consider the
individual metrics, the supporting evidence, confidence, and the distance to relevant
thresholds.

\subsection{Analysis Layers and Presentation of Results}
\label{sec:analyseschichten}

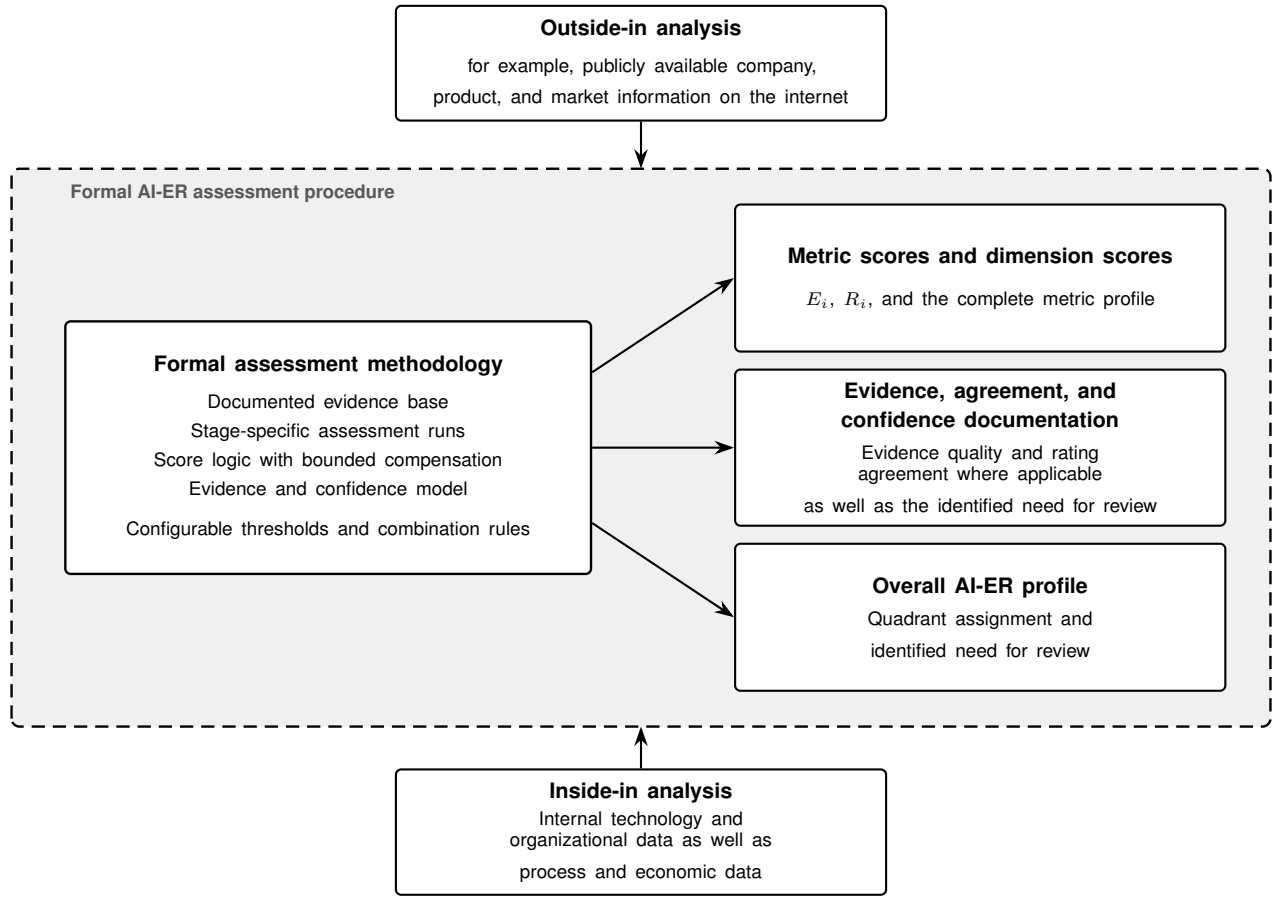
\begin{figure*}[!t]
\centering
\resizebox{0.92\textwidth}{!}{%
\begin{tikzpicture}[
  font=\sffamily,
  source/.style={
    draw=black,
    fill=white,
    line width=0.8pt,
    rounded corners=2pt,
    align=center,
    text width=5.8cm,
    minimum height=1.32cm,
    inner xsep=10pt,
    inner ysep=5pt
  },
  method/.style={
    draw=black,
    fill=white,
    line width=0.9pt,
    rounded corners=2pt,
    align=center,
    text width=6.2cm,
    minimum height=3.35cm,
    inner xsep=11pt,
    inner ysep=7pt
  },
  result/.style={
    draw=black,
    fill=white,
    line width=0.8pt,
    rounded corners=2pt,
    align=center,
    text width=5.8cm,
    minimum height=1.95cm,
    inner xsep=10pt,
    inner ysep=5pt
  },
  arrow/.style={
    -{Stealth[length=2.6mm,width=1.8mm]},
    draw=black,
    line width=0.8pt
  }
]

\node[source] (outside) at (7.75,5.10) {
  {\bfseries\footnotesize Outside-in analysis}\\[2pt]
  {\scriptsize
    for example, publicly available company, product, and market information
    on the internet
  }
};

\node[source] (inside) at (7.75,-5.10) {
  {\bfseries\footnotesize Inside-in analysis}\\[2pt]
  {\scriptsize
    Internal technology and organizational data as well as\\
    process and economic data
  }
};

\draw[
  draw=black,
  fill=black!6,
  line width=0.9pt,
  rounded corners=3pt,
  dash pattern=on 5pt off 3pt
]
  (-0.60,-3.70) rectangle (16.10,3.70);

\node[
  anchor=north west,
  fill=black!6,
  text=black!65,
  inner xsep=5pt,
  inner ysep=2pt,
  font=\bfseries\scriptsize\sffamily
]
  at (0.00,3.56)
  {Formal AI-ER assessment procedure};

\node[method] (method) at (3.60,0) {
  {\bfseries\footnotesize Formal assessment methodology}\\[6pt]
  {\scriptsize
    Documented evidence base\\[3pt]
    Stage-specific assessment runs\\[3pt]
    Score logic with bounded compensation\\[3pt]
    Evidence and confidence model\\[3pt]
    Configurable thresholds and combination rules
  }
};

\node[result] (scores) at (12.25,2.25) {
  {\bfseries\footnotesize Metric scores and dimension scores}\\[4pt]
  {\scriptsize
    $E_i$, $R_i$, and the complete metric profile
  }
};

\node[result] (confidence) at (12.25,0) {
  {\bfseries\footnotesize Evidence, agreement, and}\\[-1pt]
  {\bfseries\footnotesize confidence documentation}\\[4pt]
  {\scriptsize
    Evidence quality and rating agreement where applicable\\
    as well as the identified need for review
  }
};

\node[result] (profile) at (12.25,-2.25) {
  {\bfseries\footnotesize Overall AI-ER profile}\\[4pt]
  {\scriptsize
    Quadrant assignment and\\
    identified need for review
  }
};

\draw[arrow]
  (outside.south)
  --
  (outside.south |- 0,3.70);

\draw[arrow]
  (inside.north)
  --
  (inside.north |- 0,-3.70);

\draw[arrow]
  ([yshift=1.00cm]method.east)
  --
  (scores.west);

\draw[arrow]
  (method.east)
  --
  (confidence.west);

\draw[arrow]
  ([yshift=-1.00cm]method.east)
  --
  (profile.west);

\end{tikzpicture}%
}
\caption{Formal AI-ER methodology with outside-in and inside-in evidence and the assessment results derived from them.}
\label{fig:analyseschichten}
\end{figure*}

The assessment can be performed in two stages. An outside-in analysis uses publicly
available information to produce an initial rating of the nine metrics and the business
model modulator. It also identifies weakly supported ratings and information gaps. This
makes an initial AI-ER profile available before internal data collection is complete.
Figure~\ref{fig:analyseschichten} shows the relationship between both analysis layers and
the formal assessment methodology.

The inside-in analysis is conducted as a single subsequent assessment run. It takes
the preliminary outside-in profile as its starting point and adds internal technical,
organizational, process, and economic information. It tests assumptions formed from
public evidence and can change both metric scores and confidence. Detailed review should
focus on metrics that influence a dimension score or remain weakly supported. Ratings
close to thresholds also deserve particular attention. Both layers use the same metrics
and score logic, so the inside-in analysis refines the provisional profile rather than
replacing it with a separate assessment.

For company $i$ and metric $j$, let $x_{ij}^{O}$ denote the preliminary outside-in
rating and $x_{ij}^{I}$ the rating after the single inside-in run. Let $E_{ij}^{O}$
denote the set of public evidence and $E_{ij}^{I}$ the set of additional internal
evidence. The inside-in rating is based on $E_{ij}^{O}\cup E_{ij}^{I}$ and the same scale anchors and
score logic. The change from the preliminary rating is

\begin{equation}
\Delta x_{ij}
=
x_{ij}^{I}-x_{ij}^{O}.
\label{eq:inside-in-update}
\end{equation}

A value of $\Delta x_{ij}=0$ confirms the preliminary rating. A nonzero value indicates
a revision based on the additional internal evidence. Even when the metric score remains
unchanged, the added evidence can change $Q_{ij}$ and therefore confidence. As the
inside-in stage uses one run, $L=1$ for this stage and
Equation~(\ref{eq:metric-confidence}) reduces to $C_{ij}=Q_{ij}$.

The result consists of the two dimension scores, the quadrant position, and the
complete metric profile. Each metric is accompanied by its justification, evidence
quality, rating agreement where applicable, and confidence. Material assumptions and
remaining information gaps are recorded separately. The condensed profile therefore
remains traceable to its underlying findings.

\subsection{Illustrative Numerical Example}
\label{sec:beispiel}

A fictitious company $A$ illustrates the score logic from
Section~\ref{sec:score-logik} and the confidence model from
Section~\ref{sec:konfidenz}. The values are hypothetical and have no empirical meaning.
Three independent assessment runs are assumed for every metric, so $L=3$.
Equation~(\ref{eq:metric-median}) therefore returns an observed integer rating as the
combined metric score.
Table~\ref{tab:beispiel-exposure} shows the assumed exposure ratings and the business
model modulator.

\begin{table}[!t]
\centering
\small
\caption{Assumed ratings for the AI exposure metrics of the fictitious company $A$
(illustrative example).}
\label{tab:beispiel-exposure}
\vspace{2pt}
\begin{tabular}{lccc}
\toprule
\textbf{Metric} & \textbf{Runs $x_{A j}^{(1..3)}$} & \textbf{Median $\widetilde{x}_{Aj}$} \\
\midrule
$x_A^{\mathrm{sub}}$ & 4, 4, 5 & 4 \\
$x_A^{\mathrm{rep}}$ & 2, 3, 3 & 3 \\
$x_A^{\mathrm{comp}}$ & 4, 4, 5 & 4 \\
$x_A^{\mathrm{cust}}$ & 3, 3, 4 & 3 \\
\midrule
\multicolumn{3}{l}{$M_A^{\mathrm{bm}}$ has an amplifying effect and equals $+1$.} \\
\bottomrule
\end{tabular}
\end{table}

The core exposure drivers are $x_A^{\mathrm{sub}}=4$ and
$x_A^{\mathrm{rep}}=3$. Their maximum gives a base value of 4. Since
$x_A^{\mathrm{comp}}=4$, the exposure indicator in
Equation~(\ref{eq:exposure-indicator}) equals $I_A^E=1$. The business model modulator
equals $M_A^{\mathrm{bm}}=+1$. Equation~(\ref{eq:exposure-score}) therefore gives
\[
E_A = \operatorname{clamp}_{[1,5]}\bigl(\max\{4,3\} + 1 + 1\bigr) = \operatorname{clamp}_{[1,5]}(6) = 5.
\]

Company $A$ reaches the highest exposure level. The core benefit is highly substitutable
and competitive dynamics already reach the indicator threshold. The amplifying business
model effect increases the score by a further step.

Table~\ref{tab:beispiel-resilience} shows the assumed resilience ratings.

\begin{table}[!t]
\centering
\small
\caption{Assumed ratings for the AI resilience metrics of the fictitious company $A$
(illustrative example).}
\label{tab:beispiel-resilience}
\vspace{2pt}
\begin{tabular}{lcc}
\toprule
\textbf{Metric} & \textbf{Runs $x_{Aj}^{(1..3)}$} & \textbf{Median $\widetilde{x}_{Aj}$} \\
\midrule
$x_A^{\mathrm{prot}}$ & 3, 4, 4 & 4 \\
$x_A^{\mathrm{adapt}}$ & 3, 3, 4 & 3 \\
$x_A^{\mathrm{tech}}$ & 4, 4, 4 & 4 \\
$x_A^{\mathrm{impl}}$ & 3, 4, 4 & 4 \\
$x_A^{\mathrm{econ}}$ & 3, 3, 4 & 3 \\
\bottomrule
\end{tabular}
\end{table}

For resilience, $x_A^{\mathrm{prot}}=4$ and $x_A^{\mathrm{adapt}}=3$ are the core
drivers. Their minimum gives a base value of 3. Both technical AI maturity and
implementation capability reach 4, so $I_A^R=1$. Economic viability remains above the
penalty threshold and $M_A^{\mathrm{econ}}=0$. The resilience score is
\[
R_A = \operatorname{clamp}_{[1,5]}\bigl(\min\{4,3\} + 1 - 0\bigr) = \operatorname{clamp}_{[1,5]}(4) = 4.
\]

The company therefore combines high exposure with high resilience. Its adaptability
score limits resilience more strongly than technical AI maturity.

The confidence calculation can be illustrated with
$x_A^{\mathrm{sub}}$. Assume evidence component scores of
$q_A^{\mathrm{dir}}=1$, $q_A^{\mathrm{tim}}=1$, $q_A^{\mathrm{cmp}}=0.5$,
$q_A^{\mathrm{ind}}=1$, and $q_A^{\mathrm{agr}}=0.5$, with equal weights. Evidence
quality is $Q_{A,\mathrm{sub}}=0.8$. For the ratings $4,4,5$ with median 4, the summed
absolute deviation is 1. Equation~(\ref{eq:rating-consistency}) therefore gives
$A_{A,\mathrm{sub}}\approx0.83$. Metric confidence is the lower value and thus equals
$C_{A,\mathrm{sub}}=0.8$. The remaining decision-relevant metrics are treated in the
same way. Dimension confidence then follows from
Equation~(\ref{eq:dimension-confidence}).

With $E_A=5$ and $R_A=4$, company $A$ belongs to the
\emph{AI-Ready Compounder} quadrant in Table~\ref{tab:quadrant-thresholds}. Exposure is
high, but the company also has substantial capacity to respond. The example illustrates
why the two dimensions should not be collapsed into one score.

\section{Validation and Limits}
\label{sec:validierung}

The proposed rating logic makes AI-ER operational, but the framework has not yet been
validated empirically. Validation must address the suitability of the metrics, the
reliability of their application, the stability of the score logic, and the relationship
between assessment results and later company developments. The confidence model requires
separate examination because it is intended to distinguish well-supported ratings from
provisional ones.

\subsection{Approach to Empirical Validation}
\label{sec:validierungsansatz}

A first validation step should use comparative case studies across different
software-based business models. The cases should cover different combinations of
exposure and resilience and should vary in size, market position, and prior AI use. Each
case can first be assessed from public information and then reassessed with internal
evidence. The comparison indicates which metrics can be judged reliably from the outside
and where internal information changes the result.

Independent assessments of the same company are required to test rating reliability.
Agreement can be measured with established statistics such as Krippendorff's alpha or
Cohen's kappa \parencite{Krippendorff2018,Cohen1960}. Large deviations would indicate
unclear concepts, weak scale anchors, or excessive interpretive freedom. Validation
should also examine whether the metrics within each dimension remain empirically
distinct and whether exposure and resilience can be separated as intended.

Longitudinal validation is needed to assess predictive relevance. High exposure should
be associated with later pressure on differentiation, pricing, margins, or customer
access. High resilience should be associated with a stronger capacity to adapt products,
processes, technical structures, or the business model. Such relationships can only be
examined by comparing earlier assessments with developments observed later.

Sensitivity analysis should vary thresholds, combination rules, weights, and modulators
within plausible ranges. Stable dimension scores would support the robustness of the
model structure. Strong changes would indicate dependence on particular parameter
choices. The confidence model should also be tested against later evidence and repeated
assessments. High-confidence ratings should prove more stable than low-confidence
ratings if the confidence model works as intended.

\subsection{Limits of the Assessment Framework}
\label{sec:grenzen}

The current scale anchors, thresholds, combination rules, evidence weights, and
modulators have not been calibrated on a large sample. The maximum rule for exposure and
the minimum rule for resilience are conceptual choices. They assume that one pronounced
attack path can determine exposure and that one weak core condition can limit
resilience. Their suitability across industries and business models remains an empirical
question. The dimension scores should therefore be interpreted as structured
classifications rather than precise measurements.

Assessment quality depends strongly on the available evidence. Public information often
describes products and visible AI activities better than internal technical or
organizational conditions. Companies with extensive external communication may therefore
appear easier to assess than more reticent companies with similar capabilities. The
confidence model makes this limitation visible but cannot replace missing evidence.

AI-ER also provides a time-bound assessment. New AI capabilities, changes in cost,
competitive offerings, regulation, or internal transformation can alter both dimensions.
Assessments should therefore be updated when material conditions change. The overall
model can be used across industries, but scale anchors and evidence requirements may
need adaptation to industry-specific value creation and regulatory conditions.

A single AI-ER assessment does not establish causality. Company performance is influenced
by many factors beyond AI. The framework therefore complements strategic, technical, and
financial analysis rather than replacing them. Dimension scores should always be read
together with the individual metrics, the supporting evidence, and confidence.

\section{Conclusion and Outlook}
\label{sec:fazit}

This paper develops \emph{Artificial Intelligence Exposure and Resilience} as a framework
for assessing how AI affects software-based business models. Its central distinction is
between AI exposure and AI resilience. Exposure describes the pressure for change.
Resilience describes the company's capacity to absorb that pressure and adapt. Keeping
both dimensions separate makes situations visible in which a company is highly exposed
but also well prepared to respond.

The metrics are derived from AI capabilities, economic impact mechanisms, and research
on organizational adaptability. The proposed score logic is non-compensatory so that a
critical attack path or a weak core condition remains visible. Evidence quality and
rating agreement are assessed separately from the metric values and are combined into a
confidence measure. The framework can therefore distinguish the assessment result from
the reliability of the evidence supporting it.

AI-ER can be applied first as an outside-in assessment and later refined with internal
evidence. Possible applications include strategic review, technology due diligence, and
investment or acquisition analysis. The framework remains a proposal that requires
empirical validation. Comparative case studies, independent assessment runs,
longitudinal observation, and sensitivity analysis are needed to test the metrics and
the score logic.

A software implementation can support evidence collection and the reproducible
application of the formal rules. Such implementation work is useful for testing
practical applicability, but it does not replace empirical validation of the framework.
The next research step is therefore to combine implementation tests with a broader case
base and to refine scale anchors, thresholds, weights, and modulators where the evidence
supports adjustment.

\section*{Note on Preparation}
Technical assistance systems were used for language revision, formal checks, and
individual editorial steps. The concept, methodology, selection of content, source
assessment, argumentation, and approval of the final version remained with the authors.

\balance
\bibliographystyle{IEEEtranN}
\bibliography{references}

@article{Kanbach2024,
  author  = {Kanbach, Dominik K. and Heiduk, Louisa and Blueher, Georg and Schreiter, Maximilian and Lahmann, Alexander},
  title   = {The {GenAI} is out of the bottle: generative artificial intelligence from a business model innovation perspective},
  journal = {Review of Managerial Science},
  year    = {2024},
  volume  = {18},
  number  = {4},
  pages   = {1189--1220},
  doi     = {10.1007/s11846-023-00696-z},
  url     = {https://link.springer.com/article/10.1007/s11846-023-00696-z}
}

@article{Jorzik2024,
  author  = {Jorzik, Philip and Klein, Sascha P. and Kanbach, Dominik K. and Kraus, Sascha},
  title   = {{AI}-driven business model innovation: A systematic review and research agenda},
  journal = {Journal of Business Research},
  year    = {2024},
  volume  = {182},
  pages   = {114764},
  doi     = {10.1016/j.jbusres.2024.114764},
  url     = {https://www.sciencedirect.com/science/article/pii/S0148296324002686}
}

@article{Eloundou2024,
  author  = {Eloundou, Tyna and Manning, Sam and Mishkin, Pamela and Rock, Daniel},
  title   = {{GPTs} are {GPTs}: Labor market impact potential of {LLMs}},
  journal = {Science},
  year    = {2024},
  volume  = {384},
  number  = {6702},
  pages   = {1306--1308},
  doi     = {10.1126/science.adj0998},
  url     = {https://www.science.org/doi/10.1126/science.adj0998}
}

@misc{OECDGenAI,
  author = {{OECD}},
  title  = {Generative {AI}},
  year   = {2026},
  url    = {https://www.oecd.org/en/topics/sub-issues/generative-ai.html},
  note   = {Accessed March 29, 2026}
}

@techreport{Bick2025,
  author      = {Bick, Alexander and Blandin, Adam and Deming, David J.},
  title       = {The rapid adoption of generative {AI}},
  institution = {National Bureau of Economic Research},
  type        = {NBER Working Paper},
  number      = {32966},
  year        = {2024},
  doi         = {10.3386/w32966},
  url         = {https://www.nber.org/papers/w32966},
  note        = {Revised February 2025}
}

@book{Forsgren2018,
  author    = {Forsgren, Nicole and Humble, Jez and Kim, Gene},
  title     = {Accelerate: The Science of Lean Software and DevOps},
  publisher = {IT Revolution Press},
  year      = {2018},
  address   = {Portland, OR},
  isbn      = {978-1942788331}
}

@inproceedings{Sculley2015,
  author    = {Sculley, D. and Holt, Gary and Golovin, Daniel and Davydov, Eugene and Phillips, Todd and Ebner, Dietmar and Chaudhary, Vinay and Young, Michael and Crespo, Jean-François and Dennison, Dan},
  title     = {Hidden technical debt in machine learning systems},
  booktitle = {Advances in Neural Information Processing Systems},
  volume    = {28},
  year      = {2015},
  publisher = {Curran Associates, Inc.},
  url       = {https://proceedings.neurips.cc/paper_files/paper/2015/hash/86df7dcfd896fcaf2674f757a2463eba-Abstract.html}
}

@article{Teece1997,
  author  = {Teece, David J. and Pisano, Gary and Shuen, Amy},
  title   = {Dynamic capabilities and strategic management},
  journal = {Strategic Management Journal},
  year    = {1997},
  volume  = {18},
  number  = {7},
  pages   = {509--533},
  doi     = {10.1002/(SICI)1097-0266(199708)18:7<509::AID-SMJ882>3.0.CO;2-Z}
}

@book{Parker2016,
  author    = {Parker, Geoffrey G. and Van Alstyne, Marshall W. and Choudary, Sangeet Paul},
  title     = {Platform Revolution: How Networked Markets Are Transforming the Economy and How to Make Them Work for You},
  publisher = {W. W. Norton \& Company},
  year      = {2016},
  address   = {New York},
  isbn      = {978-0393354355}
}

@article{Wessel2025,
  author  = {Wessel, Michael and Adam, Martin and Benlian, Alexander and Majchrzak, Ann and Thies, Ferdinand},
  title   = {Generative {AI} and its Transformative Value for Digital Platforms},
  journal = {Journal of Management Information Systems},
  year    = {2025},
  volume  = {42},
  number  = {2},
  pages   = {346--369},
  doi     = {10.1080/07421222.2025.2487315},
  url     = {https://www.jmis-web.org/articles/1706}
}

@techreport{OECDCompetition2025,
  author      = {{OECD}},
  title       = {Artificial intelligence and competitive dynamics in downstream markets},
  institution = {OECD Publishing},
  year        = {2025},
  number      = {331},
  type        = {OECD Roundtables on Competition Policy Papers},
  address     = {Paris},
  doi         = {10.1787/ccf0624a-en},
  url         = {https://www.oecd.org/en/publications/artificial-intelligence-and-competitive-dynamics-in-downstream-markets_ccf0624a-en.html}
}

@article{Duchek2020,
  author  = {Duchek, Stephanie},
  title   = {Organizational resilience: a capability-based conceptualization},
  journal = {Business Research},
  year    = {2020},
  volume  = {13},
  pages   = {215--246},
  doi     = {10.1007/s40685-019-0085-7},
  url     = {https://link.springer.com/article/10.1007/s40685-019-0085-7}
}

@article{TimLeidner2023,
  author  = {Tim, Yenni and Leidner, Dorothy E.},
  title   = {Digital Resilience: A Conceptual Framework for Information Systems Research},
  journal = {Journal of the Association for Information Systems},
  year    = {2023},
  volume  = {24},
  number  = {5},
  pages   = {1184--1198},
  doi     = {10.17705/1jais.00842},
  url     = {https://aisel.aisnet.org/jais/vol24/iss5/11/}
}

@article{Buliga2016,
  author  = {Buliga, Oana and Scheiner, Christian W. and Voigt, Kai-Ingo},
  title   = {Business model innovation and organizational resilience: towards an integrated conceptual framework},
  journal = {Journal of Business Economics},
  year    = {2016},
  volume  = {86},
  number  = {6},
  pages   = {647--670},
  doi     = {10.1007/s11573-015-0796-y},
  url     = {https://link.springer.com/article/10.1007/s11573-015-0796-y}
}

@article{Han2025,
  author  = {Han, Miaozhe and Shen, Hongchuan and Wu, Jing and Zhang, Xiaoquan Michael},
  title   = {Artificial Intelligence and Firm Resilience: Empirical Evidence from Natural Disaster Shocks},
  journal = {Information Systems Research},
  year    = {2025},
  volume  = {36},
  number  = {4},
  pages   = {2116--2133},
  doi     = {10.1287/isre.2022.0440},
  url     = {https://pubsonline.informs.org/doi/10.1287/isre.2022.0440}
}

@article{Guo2026,
  author  = {Guo, Tao and Shang, Jie and Ding, Xiaozhou},
  title   = {Send charcoal in snowy weather: Artificial intelligence and organizational resilience},
  journal = {Asia Pacific Journal of Management},
  year    = {2026},
  doi     = {10.1007/s10490-025-10107-4},
  url     = {https://link.springer.com/article/10.1007/s10490-025-10107-4},
  note    = {Online first}
}

@techreport{AIIndex2026,
  author      = {{Stanford Institute for Human-Centered Artificial Intelligence}},
  title       = {The 2026 {AI} Index Report},
  institution = {Stanford University},
  year        = {2026},
  url         = {https://hai.stanford.edu/ai-index/2026-ai-index-report},
  note        = {Accessed July 27, 2026}
}

@techreport{IASafetyReport2026,
  author      = {Bengio, Yoshua and others},
  title       = {International {AI} Safety Report 2026},
  institution = {Department for Science, Innovation and Technology},
  number      = {DSIT 2026/001},
  year        = {2026},
  url         = {https://internationalaisafetyreport.org/publication/international-ai-safety-report-2026},
  note        = {Accessed July 27, 2026}
}

@techreport{NISTGenAIProfile2024,
  author      = {Autio, Chloe and Schwartz, Reva and Dunietz, Jesse and Jain, Shomik and Stanley, Martin and Tabassi, Elham and Hall, Patrick and Roberts, Kamie},
  title       = {Artificial Intelligence Risk Management Framework: Generative Artificial Intelligence Profile},
  institution = {National Institute of Standards and Technology},
  number      = {NIST AI 600-1},
  year        = {2024},
  doi         = {10.6028/NIST.AI.600-1},
  url         = {https://doi.org/10.6028/NIST.AI.600-1}
}

@misc{EUAIAct2024,
  author = {{European Parliament and Council of the European Union}},
  title  = {Regulation ({EU}) 2024/1689 laying down harmonised rules on artificial intelligence},
  year   = {2024},
  url    = {https://eur-lex.europa.eu/eli/reg/2024/1689/oj},
  note   = {Official Journal of the European Union, 12 July 2024}
}

@techreport{OECDExposure2026,
  author      = {{OECD}},
  title       = {The {OECD} {AI} Exposure Measure: Mapping the {OECD} {AI} Capability Indicators to Occupations},
  institution = {OECD Publishing},
  series      = {OECD Artificial Intelligence Papers},
  number      = {59},
  year        = {2026},
  doi         = {10.1787/f3da0f0a-en},
  url         = {https://doi.org/10.1787/f3da0f0a-en}
}

@article{Kreuzberger2023,
  author  = {Kreuzberger, Dominik and K{\"u}hl, Niklas and Hirschl, Sebastian},
  title   = {Machine Learning Operations ({MLOps}): Overview, Definition, and Architecture},
  journal = {IEEE Access},
  year    = {2023},
  volume  = {11},
  pages   = {31866--31879},
  doi     = {10.1109/ACCESS.2023.3262138},
  url     = {https://doi.org/10.1109/ACCESS.2023.3262138}
}

@book{Roy1996,
  author    = {Roy, Bernard},
  title     = {Multicriteria Methodology for Decision Aiding},
  publisher = {Springer},
  address   = {Boston, MA},
  year      = {1996}
}

@book{Munda2008,
  author    = {Munda, Giuseppe},
  title     = {Social Multi-Criteria Evaluation for a Sustainable Economy},
  publisher = {Springer},
  address   = {Berlin, Heidelberg},
  year      = {2008}
}

@article{Cohen1960,
  author  = {Cohen, Jacob},
  title   = {A Coefficient of Agreement for Nominal Scales},
  journal = {Educational and Psychological Measurement},
  volume  = {20},
  number  = {1},
  pages   = {37--46},
  year    = {1960}
}

@book{Krippendorff2018,
  author    = {Krippendorff, Klaus},
  title     = {Content Analysis: An Introduction to Its Methodology},
  edition   = {4},
  publisher = {SAGE Publications},
  address   = {Thousand Oaks, CA},
  year      = {2018}
}

\appendices
\section{Notation and Model Parameters}
\label{app:parameter}

Table~\ref{tab:notation} summarizes the symbols used in the main text.
Table~\ref{tab:parameter} lists the configurable parameters of the AI-ER assessment
model. The default values reflect the initial configuration used in this paper. Any
deviation should be documented and compared with this configuration during validation.

\begin{table*}[!t]
\centering
\caption{Notation of the formal AI-ER assessment methodology.}
\label{tab:notation}
\small
\renewcommand{\arraystretch}{1.14}

\begin{tabularx}{\textwidth}{
@{}
>{\raggedright\arraybackslash}p{3.4cm}
>{\raggedright\arraybackslash}X
@{}
}
\toprule
\textbf{Symbol} & \textbf{Meaning} \\
\midrule

\multicolumn{2}{@{}l}{\textit{Indices and assessment runs}}\\[2pt]

$i$
&
Index of the company under consideration.
\\

$j$
&
Index of a metric within the assessment model.
\\

$\ell\in\{1,\ldots,L\}$
&
Index of an independent assessment run.
\\

$L$
&
Total number of independent assessment runs.
\\

$x_{ij}^{(\ell)}$
&
Rating of metric $j$ for company $i$ in assessment run $\ell$.
\\

$\widetilde{x}_{ij}$
&
Combined score of metric $j$ for company $i$. In the configuration used here, it is
formed as the median of the independent assessment runs.
\\

$x_{ij}^{O}$
&
Preliminary outside-in rating of metric $j$ for company $i$.
\\

$x_{ij}^{I}$
&
Rating of metric $j$ for company $i$ after the single inside-in assessment run.
\\

$E_{ij}^{O},\ E_{ij}^{I}$
&
Sets of public evidence and additional internal evidence used for metric $j$ and
company $i$.
\\

$\Delta x_{ij}$
&
Difference between the inside-in and preliminary outside-in rating.
\\

\midrule
\multicolumn{2}{@{}l}{\textit{Exposure metrics and business model modulation}}\\[2pt]

$x_i^{\mathrm{sub}}$
&
Rating of the substitutability of the core benefit of company $i$.
\\

$x_i^{\mathrm{rep}}$
&
Rating of the replicability of the offering of company $i$.
\\

$x_i^{\mathrm{comp}}$
&
Rating of the competitive dynamics influenced by artificial intelligence.
\\

$x_i^{\mathrm{cust}}$
&
Rating of customer access and demand pressure.
\\

$M_i^{\mathrm{bm}}$
&
Business model modulator for company $i$ with a dampening, neutral, or amplifying effect
on the exposure assessment.
\\

\midrule
\multicolumn{2}{@{}l}{\textit{Resilience metrics and economic modulation}}\\[2pt]

$x_i^{\mathrm{prot}}$
&
Rating of the protective positions of company $i$, for example due to hard-to-imitate
resources, data, relationships, or institutional advantages.
\\

$x_i^{\mathrm{adapt}}$
&
Rating of organizational and strategic adaptability.
\\

$x_i^{\mathrm{tech}}$
&
Rating of technical AI maturity.
\\

$x_i^{\mathrm{impl}}$
&
Rating of the ability to implement AI-related initiatives organizationally and
operationally.
\\

$x_i^{\mathrm{econ}}$
&
Rating of the economic viability of AI-related investments and transformation measures.
\\

$M_i^{\mathrm{econ}}$
&
Modulator for low economic viability of company $i$ within the resilience assessment.
\\

\midrule
\multicolumn{2}{@{}l}{\textit{Indicator function and dimension scores}}\\[2pt]

$I(P),\ I_i^E,\ I_i^R$
&
Indicator function for a condition $P$ and the binary indicator values derived from it
for the exposure and resilience assessments.
\\

$E_i$
&
AI exposure dimension score for company $i$.
\\

$R_i$
&
AI resilience dimension score for company $i$.
\\

\midrule
\multicolumn{2}{@{}l}{\textit{Evidence, agreement, and confidence}}\\[2pt]

$Q_{ij}$
&
Weighted evidence quality of the information base used for metric $j$ and company $i$.
\\

$\mathcal{K}$
&
Set of the five evidence properties, with
$\mathcal{K}=
\{\mathrm{dir},\mathrm{tim},\mathrm{cmp},\mathrm{ind},\mathrm{agr}\}$
for directness, timeliness, completeness, independence, and substantive agreement.
\\

$q_{ij}^{(k)}$
&
Component score of evidence quality for company $i$ and metric $j$ for evidence property
$k\in\mathcal{K}$. Each component score lies in $\{0,0.5,1\}$.
\\

$w_j^{(k)}$
&
Positive, normalized, and optionally metric-specific weight of the evidence component
score $q_{ij}^{(k)}$ for $k\in\mathcal{K}$. The weights satisfy
$\sum_{k\in\mathcal{K}}w_j^{(k)}=1$.
\\

$A_{ij}$
&
Agreement of the independent ratings for metric $j$ and company $i$ when $L\geq 2$.
\\

$C_{ij}$
&
Confidence of the combined metric score $\widetilde{x}_{ij}$.
\\

$\mathcal{A}_i^{\mathrm{E}}$
&
Set of metrics that are decision-relevant for company $i$ within AI exposure.
\\

$\mathcal{A}_i^{\mathrm{R}}$
&
Set of metrics that are decision-relevant for company $i$ within AI resilience.
\\

$C_i^{\mathrm{E}}$
&
Confidence of the AI exposure dimension score for company $i$.
\\

$C_i^{\mathrm{R}}$
&
Confidence of the AI resilience dimension score for company $i$.
\\

\bottomrule
\end{tabularx}
\end{table*}

\begin{table*}[!t]
\centering
\caption{Configurable parameters of the AI-ER model with value ranges and default
configuration.}
\label{tab:parameter}
\vspace{2pt}

\footnotesize
\setlength{\tabcolsep}{3pt}
\renewcommand{\arraystretch}{1.16}

\begin{adjustbox}{
max width=\textwidth,
max totalheight=0.94\textheight,
keepaspectratio,
center
}
\begin{tabularx}{\textwidth}{
@{}
>{\RaggedRight\arraybackslash}p{3.7cm}
>{\RaggedRight\arraybackslash}X
>{\RaggedRight\arraybackslash}p{3.7cm}
>{\RaggedRight\arraybackslash}p{3.7cm}
@{}
}
\toprule
\textbf{Parameter}
&
\textbf{Description}
&
\textbf{Value range or options}
&
\textbf{Default value}
\\
\midrule

Rating scale
&
Ordinal metric scale with substantively defined scale anchors.
&
Anchored integer levels.
&
$\{1,\ldots,5\}$ with substantively defined scale anchors at 1, 3, and 5.
\\

Runs $L$
&
Number of independent ratings per metric.
&
$L\in\mathbb{N}$, $L\geq 1$.
&
$L=3$ for the initial outside-in assessment. The subsequent inside-in assessment
uses $L=1$. $A_{ij}$ is determined for $L\geq 2$.
\\

Run aggregation
&
Formation of $\widetilde{x}_{ij}$.
&
Median, mean, trimmed mean.
&
Median.
\\

Exposure core-driver rule
&
Combination of $x_i^{\mathrm{sub}}$ and $x_i^{\mathrm{rep}}$.
&
Maximum, minimum, weighted mean.
&
Maximum.
\\

Exposure indicator threshold
&
High level of $x_i^{\mathrm{comp}}$ or $x_i^{\mathrm{cust}}$.
&
$\{2,\ldots,5\}$.
&
$4$.
\\

Exposure indicator logic
&
Combination of $x_i^{\mathrm{comp}}$ and $x_i^{\mathrm{cust}}$.
&
Disjunction, conjunction.
&
Disjunction.
\\

Exposure indicator increment
&
Increase of the exposure base value when the condition is fulfilled.
&
$\{0,1,2\}$.
&
$+1$.
\\

Business model modulator
&
Dampening, neutral, or amplifying effect of the business model on the exposure score.
&
$\{-1,0,+1\}$.
&
$-1$, $0$, or $+1$ according to the categorical rating.
\\

Resilience core-driver rule
&
Combination of $x_i^{\mathrm{prot}}$ and $x_i^{\mathrm{adapt}}$.
&
Minimum, maximum, weighted mean.
&
Minimum.
\\

Resilience indicator threshold
&
High level of $x_i^{\mathrm{tech}}$ and $x_i^{\mathrm{impl}}$.
&
$\{2,\ldots,5\}$.
&
$4$.
\\

Resilience indicator logic
&
Combination of $x_i^{\mathrm{tech}}$ and $x_i^{\mathrm{impl}}$.
&
Conjunction, disjunction.
&
Conjunction.
\\

Resilience indicator increment
&
Increase of the resilience base value when the condition is fulfilled.
&
$\{0,1,2\}$.
&
$+1$.
\\

Economic viability trigger
&
Activation of $M_i^{\mathrm{econ}}$ for low economic viability.
&
Threshold from $\{1,2,3\}$.
&
$x_i^{\mathrm{econ}}\leq 2$.
\\

Economic viability penalty
&
Reduction of the resilience score when $M_i^{\mathrm{econ}}$ is active. As defined in
Equation~(\ref{eq:economic-modulator}), $M_i^{\mathrm{econ}}$ is itself an indicator and
therefore fixed at $1$ when active. A different fixed penalty would require generalizing
Equation~(\ref{eq:economic-modulator}) to $M_i^{\mathrm{econ}} = p^{\mathrm{econ}}\cdot
I(x_i^{\mathrm{econ}}\leq 2)$ with an explicit weight $p^{\mathrm{econ}}$.
&
$\{1\}$ under the definition used here. An earlier unpublished prototype used an
additive penalty of $2$ under a different (non-indicator) definition of
$M_i^{\mathrm{econ}}$.
&
$1$.
\\

Dimension range
&
Value range of $E_i$ and $R_i$.
&
Scale-dependent.
&
$[1,5]$.
\\

Evidence grading
&
Values of the evidence component scores $q_{ij}^{(k)}$.
&
Discrete levels.
&
$\{0,0.5,1\}$.
\\

Evidence weights $w_j^{(k)}$
&
Relative weighting of the five evidence properties directness, timeliness, completeness,
independence, and substantive agreement in the calculation of $Q_{ij}$. The weights may
be set per metric or uniformly across metrics.
&
$w_j^{(k)}>0$ for
$k\in\{\mathrm{dir},\mathrm{tim},\mathrm{cmp},\mathrm{ind},\mathrm{agr}\}$ with
$\sum_{k\in\mathcal{K}}w_j^{(k)}=1$.
&
Uniform weights for all metrics and equal weighting of the five evidence properties.
\\

Evidence aggregation
&
Formation of $Q_{ij}$ from the evidence component scores and their associated weights.
&
Weighted mean with normalized weights.
&
$\displaystyle
Q_{ij}
=
\sum_{k\in\mathcal{K}} w_j^{(k)}q_{ij}^{(k)}$.
\\

Rating agreement
&
Deviations of the individual ratings from the median.
&
Normalized absolute deviation.
&
Normalization by $2L$.
\\

Confidence combination
&
Combination of $Q_{ij}$ and $A_{ij}$.
&
Minimum, product, weighted mean.
&
$\min\!\{Q_{ij},A_{ij}\}$ for $L\geq 2$ and $Q_{ij}$ for $L=1$.
\\

Confidence thresholds
&
Low, medium, and high confidence.
&
$0<\theta_1<\theta_2<1$.
&
$\theta_1=0.5$, $\theta_2=0.75$.
\\

Dimension confidence
&
Aggregation over $\mathcal{A}_i^{\mathrm{E}}$ and
$\mathcal{A}_i^{\mathrm{R}}$.
&
Minimum, mean.
&
Minimum.
\\

Quadrant threshold
&
Separation of low and high dimension scores.
&
Between adjacent scale levels.
&
High from 4.
\\

Treatment of score 3
&
Treatment of medium dimension scores.
&
Flag or transitional category.
&
Near-threshold flag.
\\

\bottomrule
\end{tabularx}
\end{adjustbox}
\end{table*}

\end{document}